\documentclass[11pt,a4paper]{article}
\usepackage[hyperref]{acl2019}
\usepackage{colortbl}
\usepackage{times}
\usepackage{tikz}
\usepackage{bbold}

\newif\ifcomment\commentfalse

\usepackage{framed}
\usepackage{mdwlist}
\usepackage{latexsym}
\usepackage{xcolor}
\usepackage{nicefrac}
\usepackage{booktabs}
\usepackage{amsfonts}
\usepackage[T1]{fontenc}
\usepackage{bold-extra}
\usepackage{amsmath}
\usepackage{dsfont}
\usepackage{amssymb}
\usepackage{bm}
\usepackage{graphicx}
\usepackage{mathtools}
\usepackage{microtype}
\usepackage{multirow}
\usepackage{multicol}
\usepackage{latexsym,comment}

\newcommand{\gem}[1]{\mbox{\textsc{gem}}}
\newcommand{\abr}[1]{\textsc{#1}}

\newcommand{\hidetext}[1]{}
\newcommand{\ignore}[1]{}

\ifcomment
\newcommand{\pinaforecomment}[3]{\colorbox{#1}{\parbox{.8\linewidth}{#2: #3}}}
\else
\newcommand{\pinaforecomment}[3]{}
\fi

\newcommand{\smallurl}[1]{ \begin{tiny}\url{#1}\end{tiny}}

\definecolor{lightblue}{HTML}{3cc7ea}
\definecolor{CUgold}{HTML}{CFB87C}
\definecolor{grey}{rgb}{0.95,0.95,0.95}
\definecolor{ceil}{rgb}{0.57, 0.63, 0.81}

\newcommand{\squad}{\textsc{sq}{\small u}\textsc{ad}}

\aclfinalcopy
\def\aclpaperid{513}

\newcommand{\bert}{\textsc{bert}}

\newcommand{\snli}{\textsc{snli}}
\newcommand{\vqa}{\textsc{vqa}}

\definecolor{lightgrey}{rgb}{0.85,0.85,0.85}

\title{Misleading Failures of Partial-input Baselines}

\author{Shi Feng \\
Computer Science \\
University of Maryland \\
\texttt{shifeng@umiacs.umd.edu} \And
Eric Wallace \\
Allen Institute for \\
Artificial Intelligence \\
\texttt{ericw@allenai.org} \And
Jordan Boyd-Graber \\
Computer Science, iSchool, \\
\abr{umiacs}, and \abr{lsc} \\
University of Maryland\\
\texttt{jbg@umiacs.umd.edu} \\
}

\date{}

\begin{document}

\maketitle

\begin{abstract}
Recent work establishes dataset difficulty and removes annotation artifacts via
partial-input baselines (e.g., hypothesis-only models for \snli{} or question-only
models for \vqa{}).
When a partial-input baseline gets high accuracy, a dataset is cheatable.
However, the converse is not necessarily true: the failure of a partial-input baseline does
not mean a dataset is free of artifacts.
To illustrate this, we first design artificial datasets which contain trivial patterns in the full input that are undetectable by any partial-input model.
Next, we identify such artifacts in the \snli{} dataset---a hypothesis-only model
augmented with trivial patterns in the premise can solve 15\% of
the examples that are previously considered ``hard''. 
Our work provides a caveat for the use of partial-input baselines for dataset
verification and creation.
\end{abstract}
\section{Dataset Artifacts Hurt Generalizability}
\label{sec:intro}

Dataset quality is crucial for the development and evaluation of machine
learning models.
Large-scale natural language processing (\abr{nlp})
datasets often use human annotations on web-crawled data,
which can introduce \emph{artifacts}.
For example, crowdworkers might use specific words to contradict a
given premise~\cite{gururangan2018annotation}.
These artifacts corrupt the intention of the datasets to train and evaluate models for
natural language understanding.
Importantly, a human inspection of individual examples cannot catch artifacts because they
are only visible in aggregate on the dataset level.
However, machine learning algorithms, which detect and exploit
recurring patterns in large datasets by design, can just as easily use
artifacts as real linguistic clues.
As a result, models trained on these datasets can
achieve high test accuracy by exploiting artifacts but
fail to generalize, e.g., they fail under adversarial
evaluation~\cite{jia2017adversarial, ribeiro2018semantically}.

The identification of dataset artifacts has changed model evaluation and
dataset construction~\cite{chen2016thorough, jia2017adversarial,
goyal2017vqa2}.
One key method is to use partial-input baselines, i.e., models that
intentionally ignore portions of the input.
Example use cases include hypothesis-only
models for natural language inference~\cite{gururangan2018annotation},
question-only models for visual question answering~\cite{goyal2017vqa2},
and paragraph-only models for reading comprehension~\cite{kaushik2018reading}.
A successful partial-input baseline indicates that a dataset contains
artifacts which make it easier than expected.
On the other hand, examples where this baseline fails are
``hard''~\cite{gururangan2018annotation}, and the failure of
partial-input baselines is considered a verdict of a dataset's
difficulty~\cite{zellers2018swag, kaushik2018reading}.

These partial-input analyses are valuable and indeed reveal
dataset issues; however, they do not tell the whole story.
Just as being free of one ailment is not the same as a clean bill of
health, a baseline's failure only indicates that a dataset is not broken
in one specific way.
There is no reason that artifacts only infect part of the
input---models can exploit patterns that are only
visible in the full input.

After reviewing partial-input baselines
(Section~\ref{sec:partial}), we construct variants of a natural
language inference dataset to highlight the potential pitfalls of
partial-input dataset validation (Section~\ref{sec:datasets}).
Section~\ref{sec:heuristics} shows that real datasets have
artifacts that evade partial-input baselines; we use a
hypothesis-plus-one-word model to solve 15\% of the ``hard''
examples from \snli~\cite{bowman2015snli, gururangan2018annotation}
where hypothesis-only models fail. Furthermore, we highlight some of the artifacts
learned by this model using $k$-nearest neighbors in
representation space.
Section~\ref{sec:discussion} discusses how partial-input baselines should
be used in future dataset creation and analysis.

\section{What are Partial-input Baselines?}
\label{sec:partial}

A long-term goal of \abr{nlp} is to solve tasks that we
believe require a human-level understanding of language.
The \abr{nlp} community typically defines tasks with datasets: reproduce
these answers given these inputs, and you have solved the underlying
task. This task-dataset equivalence is only valid when the dataset accurately
represents the task. Unfortunately, verifying this equivalence via humans is
fundamentally insufficient: humans reason about examples one by one,
while models can discover recurring patterns.
Patterns that are not part of the underlying task, or \emph{artifacts} of
the data collection process, can lead to models that ``cheat''---ones that
achieve high test accuracy using patterns that do not
generalize.

One frequent type of artifact, especially in
classification datasets where each input contains multiple parts (e.g., a
question and an image), is a strong correlation between a part of the
input and the label.
For example, a model can answer many \abr{vqa} questions without
looking at the image~\cite{goyal2017vqa2}.
These artifacts can be detected using partial-input baselines: models
that are restricted to using only part of the input. Validating a dataset
with a partial-input baseline has the following steps:
\begin{enumerate*}
\item Decide which part of the input to use.
\item Reduce all examples in the training set and the test set.
\item Train a new model from scratch on the partial-input training
    set.
\item Test the model on the partial-input test set.
\end{enumerate*}

High accuracy from a partial-input model implies the \emph{original}
dataset is solvable (to some extent) in the wrong ways, i.e., using
unintended patterns.
Partial-input baselines have identified artifacts in many datasets, e.g.,
\abr{snli}~\cite{gururangan2018annotation, poliak2018hypothesis},
\abr{vqa}~\cite{goyal2017vqa2},
EmbodiedQA~\cite{anand2018blindfold},
visual dialogue~\cite{massiceti2018without},
and visual navigation~\cite{thomason2018baseline}.

\section{How Partial-input Baselines Fail}
\label{sec:datasets}

If a partial-input baseline fails, e.g., it gets close to
chance accuracy, one might conclude that a dataset is difficult.
For example, partial-input baselines are used to identify the ``hard''
examples in \snli{}~\cite{gururangan2018annotation}, 
verify that \squad{} is well constructed~\cite{kaushik2018reading},
and that \abr{swag} is challenging~\cite{zellers2018swag}.

Reasonable as it might seem, this kind of argument can be misleading---it is
important to understand what exactly these results do and do not
imply. A low accuracy from a partial-input baseline only means that the model
failed to confirm a specific exploitable pattern in the part of the input that the
model can see.
This does not mean, however, that the dataset is free of
artifacts---the full input might still contain very trivial patterns.

To illustrate how the failures of partial-input baselines might shadow
more trivial patterns that are only visible in the full input,
we construct two variants of the \snli{}
dataset~\cite{bowman2015snli}.
The datasets are constructed to contain trivial patterns that partial-input
baselines cannot exploit, i.e., the patterns are only visible in the full input.
As a result, a full-input can achieve perfect accuracy whereas partial-input
models fail.

\subsection{Label as Premise}

In \snli{},
each example consists of a pair of sentences: a premise and a
hypothesis. The goal is to classify the semantic relationship between
the premise and the hypothesis---either entailment, neutral, or
contradiction. 

Our first \snli{} variant is an extreme example of artifacts
that cannot be detected by a hypothesis-only
baseline.
Each \snli{} example (training and testing) is copied three times,
and the copies are assigned the labels Entailment, Neutral,
and Contradiction, respectively.  
We then set each example's premise to be the literal word of the associated
label: ``Entailment'', ``Neutral'', or ``Contradiction''
(Table~\ref{table:dataset1}).
From the perspective of a hypothesis-only model, the three copies have
identical inputs but conflicting labels. 
Thus, the best accuracy from any hypothesis-only model is
chance---the model fails due to high Bayes error.
However, a full-input model can see the label in the
premise and achieve perfect accuracy.

This serves as an extreme example of a dataset that passes a
partial-input baseline test but still contains artifacts.
Obviously, a premise-only baseline can detect these artifacts; we
address this in the next dataset variant.

\begin{table}[t]
\centering
{
\begin{tabular}{ll}
\textbf{Old Premise} & Animals are running \\
\textbf{New Premise} & Entailment \\
\textbf{Hypothesis} & Animals are outdoors \\\midrule
\textbf{Label} & Entailment
\end{tabular}
}
\caption{Each example in this dataset has the ground-truth label set as the
premise. Every hypothesis occurs three times in the dataset,
each time with a unique label and premise combination (not shown in this table).
Therefore, a hypothesis-only baseline will only achieve chance accuracy, but a
full-input model can trivially solve the dataset.}
\label{table:dataset1}
\end{table}

\subsection{Label Hidden in Premise and Hypothesis}

The artifact we introduce in the previous dataset can be easily
detected by a premise-only baseline. In this variant, we ``encrypt''
the label such that it is only visible if we combine the premise and
the hypothesis, i.e., neither premise-only nor hypothesis-only baselines
can detect the artifact.
Each label is represented by the concatenation of two ``code words'', and
this mapping is one-to-many: each label has three combinations of code words, and
each combination uniquely identifies a label.
Table~\ref{table:codewords} shows our code word configuration.
The design of the code words ensures that
a single code word cannot uniquely identify a label---you need both.

We put one code word in the premise and the other in the hypothesis. 
These encrypted labels mimic an artifact that requires
both parts of the input.
Table~\ref{table:codewords-example} shows an \snli{} example modified accordingly.
A full-input model can exploit the artifact
and trivially achieve perfect accuracy, but a partial-input model cannot.

A more extreme version of this modified dataset has exactly the nine
combinations in Table~\ref{table:codewords} as both the training set and the
test set.
Since a single code word cannot identify the label, neither
hypothesis-only nor premise-only baselines can achieve more than chance
accuracy. However, a full-input model can perfectly extract the
label by combining the premise and the hypothesis.

\begin{table}[t]
\centering
\begin{tabular}{lccc}
\textbf{Label} & \multicolumn{3}{c}{\textbf{Combinations}} \\\midrule
Entailment    & $\mathbb{A}$+$\mathbb{B}$ & $\mathbb{C}$+$\mathbb{D}$ & $\mathbb{E}$+$\mathbb{F}$ \\
Contradiction & $\mathbb{A}$+$\mathbb{F}$ & $\mathbb{C}$+$\mathbb{B}$ & $\mathbb{E}$+$\mathbb{D}$ \\
Neutral       & $\mathbb{A}$+$\mathbb{D}$ & $\mathbb{C}$+$\mathbb{F}$ & $\mathbb{E}$+$\mathbb{B}$
\end{tabular}
\caption{We ``encrypt'' the labels to mimic an artifact that requires
  both parts of the input. Each capital letter is a code word, and
  each label is derived from the combination of two
  code words. Each
  combination uniquely identifies a label, e.g., $\mathbb{A}$ in the
  premise and $\mathbb{B}$ in the hypothesis equals Entailment. However, a
  single code word cannot identify the label.}
\label{table:codewords}
\end{table}

\begin{table}[t]
\centering
{
\begin{tabular}{ll}
\textbf{Premise} & $\mathbb{A}$ Animals are running \\
\textbf{Hypothesis} & $\mathbb{B}$ Animals are outdoors \\\midrule
\textbf{Label} & Entailment
\end{tabular}
}
\caption{Each example in this dataset has a code word added to both the premise and
the hypothesis. Following the configuration of Table~\ref{table:codewords},
$\mathbb{A}$ in the premise combined with $\mathbb{B}$ in the hypothesis
indicates the label is Entailment. A full-input model can easily exploit this
artifact but partial-input models cannot.}
\label{table:codewords-example}
\end{table}
\section{Artifacts Evade Partial-input Baselines}
\label{sec:heuristics}

Our synthetic dataset variants contain trivial artifacts that partial-input
baselines fail to detect.
Do real datasets such as \snli{} have artifacts that are not
detected by partial-input baselines?

We investigate this by providing additional information about the premise to a hypothesis-only model.
In particular, we provide the last noun of the premise, i.e., we form a hypothesis-plus-one-word model.
Since this additional information appears useless to humans (examples below), it is an artifact
rather than a generalizable pattern. 

We use a \bert{}-based~\cite{devlin2018bert} classifier that gets
88.28\% accuracy with the regular, full input.
The hypothesis-only version reaches 70.10\%
accuracy.\footnote{\citet{gururangan2018annotation} report
67.0\% using a simpler hypothesis-only model.}
With the hypothesis-plus-one-word model, the accuracy improves to 74.6\%, i.e., the
model solves 15\% of the ``hard'' examples that are
unsolvable by the hypothesis-only model.\footnote{We create the easy-hard split
of the dataset using our model, not using the model from
\citet{gururangan2018annotation}.}

Table~\ref{table:plus-one-word} shows examples that are only
solvable with the one additional word from the premise.
For both the hypothesis-only and hypothesis-plus-one-word models, we follow \citet{papernot2018dknn}
and \citet{wallace2018interpreting} and retrieve training examples using
nearest neighbor search in the final \bert{} representation space. 
In the first example, humans would not consider the hypothesis ``The young boy is
crying'' as a contradiction to the premise ``camera''. In this case,
the hypothesis-only model incorrectly predicts Entailment, however, the
hypothesis-plus-one-word model
correctly predicts Contradiction.
This pattern---including one premise word---is an artifact that
regular partial-input baselines cannot detect but can be exploited by
a full-input model. 

\begin{table*}[t]
\footnotesize
\centering
\begin{tabular}{lll}
\textbf{Label} & \textbf{Premise} & \textbf{Hypothesis} \\
\toprule
\rowcolor{lightgrey}
Contradiction & A young boy hanging on a pole smiling at the \underline{camera}. & \underline{The young boy is crying.} \\
Contradiction & A boy smiles tentatively at the \underline{camera}. & \underline{a boy is crying.} \\
Contradiction & A happy child smiles at the \underline{camera}. & \underline{The child is crying at the playground.} \\
Contradiction & A girl shows a small child her \underline{camera}. & \underline{A boy crying.} \\
Entailment & A little boy with a baseball on his shirt is crying. & \underline{A boy is crying.} \\
Entailment & Young boy crying in a stroller. & \underline{A boy is crying.} \\
Entailment & A baby boy in overalls is crying. & \underline{A boy is crying.} \\
\midrule
\rowcolor{lightgrey}
Entailment & Little boy playing with his toy \underline{train}. & \underline{A boy is playing with toys.} \\
Entailment & A little boy is looking at a toy \underline{train}. & \underline{A boy is looking at a toy.} \\
Entailment & Little redheaded boy looking at a toy \underline{train}.  & \underline{A little boy is watching a toy train.} \\
Entailment & A young girl in goggles riding on a toy \underline{train}. & \underline{A girl rides a toy train.} \\
Contradiction & A little girl is playing with tinker toys. & \underline{A little boy is playing with toys.} \\
Contradiction & A toddler shovels a snowy driveway with a shovel. & \underline{A young child is playing with toys.} \\
Contradiction & A boy playing with toys in a bedroom. & \underline{A boy is playing with toys at the park.} \\
\bottomrule
\end{tabular}
\caption{We create a hypothesis-plus-one-word model that sees the hypothesis alongside the last noun in the premise. We show two \snli{} test examples (highlighted) that are answered correctly using this model but are answered incorrectly using
a hypothesis-only model. For each test example, we also show the training examples that are
nearest neighbors in \bert{}'s representation space. When using the hypothesis and the last noun in the premise (underlined), training examples with the correct label are retrieved; when using only the hypothesis,
examples with the incorrect label are retrieved.}
\label{table:plus-one-word}
\end{table*}

\section{Discussion and Related Work}
\label{sec:discussion}

Partial-input baselines are valuable sanity checks for datasets, but as we
illustrate, their implications should be understood carefully.
This section discusses methods for validating and creating datasets in light of
possible artifacts from the annotation process, as well as empirical results
that corroborate the potential pitfalls highlighted in this paper. 
Furthermore, we discuss alternative approaches for developing robust \abr{nlp}
models.

\paragraph{Hypothesis Testing}
Validating datasets with partial-input baselines is a form of
hypothesis-testing: one hypothesizes trivial solutions to the dataset (i.e.,
a spurious correlation between labels and a part of the input) and verifies if these
hypotheses are true.
While it is tempting to hypothesize other ways a model can cheat, it is
infeasible to enumerate over all of them.
In other words, if we could write down all the necessary tests for \emph{test-driven
development}~\cite{beck2002tdd} of a machine learning model, we would already
have a rule-based system that can solve our task.

\paragraph{Adversarial Annotation}
Rather than using partial-input baselines as post-hoc tests, a natural idea is
to incorporate them into the data generation process to reject bad examples. 
For example, the \abr{swag}~\cite{zellers2018swag} dataset consists of
multiple-choice answers that are selected adversarially against
an ensemble of partial-input and heuristic classifiers. 
However, since these classifiers can be
easily fooled if they rely on superficial patterns, the resulting dataset
may still contain artifacts.  In particular,
a much stronger model (\bert{}) that sees the full-input easily solves the dataset.
This demonstrates that using partial-input baselines as adversaries may lead to datasets
that are \emph{just difficult enough} to fool the baselines but not difficult
enough to ensure that no model can cheat. 

\paragraph{Adversarial Evaluation}
Instead of validating a dataset, one can alternatively
probe the model directly. 
For example, models can be stress tested using
adversarial examples~\cite{jia2017adversarial,wallace2018trick}
and challenge sets~\cite{glockner2018breaking,naik2018stress}.
These tests can reveal strikingly simple
model limitations, e.g., basic paraphrases can fool
textual entailment and
visual question answering systems~\cite{iyyer2018scpn,
ribeiro2018semantically}, while common typos drastically degrade
neural machine translation quality~\cite{belinkov2017synthetic}. 

\paragraph{Interpretations} Another technique for probing models
is to use interpretation methods. Interpretations, however,
have a problem of
faithfulness~\cite{rudin2018please}: they approximate (often locally)
a complex model 
with a simpler, interpretable model (often
a linear model). Since interpretations are inherently an
approximation, they can
never be completely faithful---there are cases where the original
model and the simple model behave differently~\cite{ghorbani2017interpretation}.
These cases might
also be especially important as they usually reflect the counter-intuitive
brittleness of the complex models (e.g., in adversarial examples).

\paragraph{Certifiable Robustness} Finally, an alternative approach for
creating models that are free of artifacts is to alter the training process.
In particular, model robustness research in computer vision
has begun to transition from
an empirical arms race between attackers and defenders to more
theoretically sound robustness
methods. For instance, convex relaxations can train
models that are provably robust 
to adversarial examples~\cite{raghunathan2018certify,wong2018provable}. 
Despite these method's impressive (and rapidly developing) results,
they largely focus on adversarial perturbations
bounded to an $L_\infty$ ball. This is due to the difficulties
in formalizing attacks and defenses for more complex threat models,
of which the discrete nature of \abr{nlp} is included. Future work can look
to generalize these
methods to other classes of model vulnerabilities and artifacts. 

\section{Conclusion}
\label{sec:conclusion}

Partial-input baselines are valuable sanity checks for dataset
difficulty, but their implications should be analyzed carefully.
We illustrate in both synthetic and real datasets how partial-input baselines
can overshadow trivial, exploitable patterns that are only visible in the
full input. Our work provides an alternative view on the use of
partial-input baselines in future dataset creation.

\section*{Acknowledgments}

This work was supported by \abr{nsf} Grant \abr{iis}-1822494.
Boyd-Graber and Feng are also supported by DARPA award
HR0011-15-C-0113 under subcontract to Raytheon BBN Technologies.  Any
opinions, findings, conclusions, or recommendations expressed here are
those of the authors and do not necessarily reflect the view of the
sponsor.

\newpage

\bibliographystyle{acl_natbib_2019}
\bibliography{journal-full,fs}

\begin{thebibliography}{25}
\expandafter\ifx\csname natexlab\endcsname\relax\def\natexlab#1{#1}\fi

\bibitem[{Anand et~al.(2018)Anand, Belilovsky, Kastner, Larochelle, and
  Courville}]{anand2018blindfold}
Ankesh Anand, Eugene Belilovsky, Kyle Kastner, Hugo Larochelle, and Aaron
  Courville. 2018.
\newblock Blindfold baselines for embodied {QA}.
\newblock In \emph{NeurIPS Visually-Grounded Interaction and Language
  Workshop}.

\bibitem[{Beck(2002)}]{beck2002tdd}
Kent Beck. 2002.
\newblock \emph{Test-Driven Development by Example}.
\newblock Addison-Wesley.

\bibitem[{Belinkov and Bisk(2018)}]{belinkov2017synthetic}
Yonatan Belinkov and Yonatan Bisk. 2018.
\newblock Synthetic and natural noise both break neural machine translation.
\newblock In \emph{Proceedings of the International Conference on Learning
  Representations}.

\bibitem[{Bowman et~al.(2015)Bowman, Angeli, Potts, and
  Manning}]{bowman2015snli}
Samuel~R. Bowman, Gabor Angeli, Christopher Potts, and Christopher~D. Manning.
  2015.
\newblock A large annotated corpus for learning natural language inference.
\newblock In \emph{Proceedings of Empirical Methods in Natural Language
  Processing}.

\bibitem[{Chen et~al.(2016)Chen, Bolton, and Manning}]{chen2016thorough}
Danqi Chen, Jason Bolton, and Christopher~D. Manning. 2016.
\newblock A thorough examination of the {CNN}/{Daily Mail} reading
  comprehension task.
\newblock In \emph{Proceedings of the Association for Computational
  Linguistics}.

\bibitem[{Devlin et~al.(2019)Devlin, Chang, Lee, and
  Toutanova}]{devlin2018bert}
Jacob Devlin, Ming-Wei Chang, Kenton Lee, and Kristina Toutanova. 2019.
\newblock {BERT}: Pre-training of deep bidirectional transformers for language
  understanding.
\newblock In \emph{Conference of the North American Chapter of the Association
  for Computational Linguistics}.

\bibitem[{Ghorbani et~al.(2019)Ghorbani, Abid, and
  Zou}]{ghorbani2017interpretation}
Amirata Ghorbani, Abubakar Abid, and James~Y. Zou. 2019.
\newblock Interpretation of neural networks is fragile.
\newblock In \emph{Association for the Advancement of Artificial Intelligence}.

\bibitem[{Glockner et~al.(2018)Glockner, Shwartz, and
  Goldberg}]{glockner2018breaking}
Max Glockner, Vered Shwartz, and Yoav Goldberg. 2018.
\newblock Breaking {NLI} systems with sentences that require simple lexical
  inferences.
\newblock In \emph{Proceedings of the Association for Computational
  Linguistics}.

\bibitem[{Goyal et~al.(2017)Goyal, Khot, Summers-Stay, Batra, and
  Parikh}]{goyal2017vqa2}
Yash Goyal, Tejas Khot, Douglas Summers-Stay, Dhruv Batra, and Devi Parikh.
  2017.
\newblock Making the {V} in {VQA} matter: Elevating the role of image
  understanding in visual question answering.
\newblock In \emph{Computer Vision and Pattern Recognition}.

\bibitem[{Gururangan et~al.(2018)Gururangan, Swayamdipta, Levy, Schwartz,
  Bowman, and Smith}]{gururangan2018annotation}
Suchin Gururangan, Swabha Swayamdipta, Omer Levy, Roy Schwartz, Samuel~R.
  Bowman, and Noah~A. Smith. 2018.
\newblock Annotation artifacts in natural language inference data.
\newblock In \emph{Conference of the North American Chapter of the Association
  for Computational Linguistics}.

\bibitem[{Iyyer et~al.(2018)Iyyer, Wieting, Gimpel, and
  Zettlemoyer}]{iyyer2018scpn}
Mohit Iyyer, John Wieting, Kevin Gimpel, and Luke~S. Zettlemoyer. 2018.
\newblock Adversarial example generation with syntactically controlled
  paraphrase networks.
\newblock In \emph{Conference of the North American Chapter of the Association
  for Computational Linguistics}.

\bibitem[{Jia and Liang(2017)}]{jia2017adversarial}
Robin Jia and Percy Liang. 2017.
\newblock Adversarial examples for evaluating reading comprehension systems.
\newblock In \emph{Proceedings of Empirical Methods in Natural Language
  Processing}.

\bibitem[{Kaushik and Lipton(2018)}]{kaushik2018reading}
Divyansh Kaushik and Zachary~C. Lipton. 2018.
\newblock How much reading does reading comprehension require? a critical
  investigation of popular benchmarks.
\newblock In \emph{Proceedings of Empirical Methods in Natural Language
  Processing}.

\bibitem[{Massiceti et~al.(2018)Massiceti, Dokania, Siddharth, and
  Torr}]{massiceti2018without}
Daniela Massiceti, Puneet~K. Dokania, N.~Siddharth, and Philip~H.S. Torr. 2018.
\newblock Visual dialogue without vision or dialogue.
\newblock In \emph{NeurIPS Workshop on Critiquing and Correcting Trends in
  Machine Learning}.

\bibitem[{Naik et~al.(2018)Naik, Ravichander, Sadeh, Rose, and
  Neubig}]{naik2018stress}
Aakanksha Naik, Abhilasha Ravichander, Norman Sadeh, Carolyn Rose, and Graham
  Neubig. 2018.
\newblock Stress test evaluation for natural language inference.
\newblock In \emph{Proceedings of International Conference on Computational
  Linguistics}.

\bibitem[{Papernot and McDaniel(2018)}]{papernot2018dknn}
Nicolas Papernot and Patrick~D. McDaniel. 2018.
\newblock Deep k-nearest neighbors: Towards confident, interpretable and robust
  deep learning.
\newblock \emph{arXiv preprint arXiv: 1803.04765}.

\bibitem[{Poliak et~al.(2018)Poliak, Naradowsky, Haldar, Rudinger, and
  Durme}]{poliak2018hypothesis}
Adam Poliak, Jason Naradowsky, Aparajita Haldar, Rachel Rudinger, and
  Benjamin~Van Durme. 2018.
\newblock Hypothesis only baselines in natural language inference.
\newblock In \emph{7th Joint Conference on Lexical and Computational Semantics
  (*SEM)}.

\bibitem[{Raghunathan et~al.(2018)Raghunathan, Steinhardt, and
  Liang}]{raghunathan2018certify}
Aditi Raghunathan, Jacob Steinhardt, and Percy Liang. 2018.
\newblock Certified defenses against adversarial examples.
\newblock In \emph{Proceedings of the International Conference on Learning
  Representations}.

\bibitem[{Ribeiro et~al.(2018)Ribeiro, Singh, and
  Guestrin}]{ribeiro2018semantically}
Marco~Tulio Ribeiro, Sameer Singh, and Carlos Guestrin. 2018.
\newblock Semantically equivalent adversarial rules for debugging {NLP} models.
\newblock In \emph{Proceedings of the Association for Computational
  Linguistics}.

\bibitem[{Rudin(2018)}]{rudin2018please}
Cynthia Rudin. 2018.
\newblock Please stop explaining black box models for high stakes decisions.
\newblock In \emph{NeurIPS Workshop on Critiquing and Correcting Trends in
  Machine Learning}.

\bibitem[{Thomason et~al.(2019)Thomason, Gordan, and
  Bisk}]{thomason2018baseline}
Jesse Thomason, Daniel Gordan, and Yonatan Bisk. 2019.
\newblock Shifting the baseline: Single modality performance on visual
  navigation \& {QA}.
\newblock In \emph{Conference of the North American Chapter of the Association
  for Computational Linguistics}.

\bibitem[{Wallace et~al.(2018)Wallace, Feng, and
  Boyd-Graber}]{wallace2018interpreting}
Eric Wallace, Shi Feng, and Jordan Boyd-Graber. 2018.
\newblock Interpreting neural networks with nearest neighbors.
\newblock In \emph{{EMNLP} Workshop {BlackboxNLP}: Analyzing and Interpreting
  Neural Networks for NLP}.

\bibitem[{Wallace et~al.(2019)Wallace, Rodriguez, Feng, Yamada, and
  Boyd-Graber}]{wallace2018trick}
Eric Wallace, Pedro Rodriguez, Shi Feng, Ikuya Yamada, and Jordan Boyd-Graber.
  2019.
\newblock Trick me if you can: Human-in-the-loop generation of adversarial
  examples for question answering.
\newblock In \emph{Transactions of the Association for Computational
  Linguistics}.

\bibitem[{Wong and Kolter(2018)}]{wong2018provable}
Eric Wong and J.~Zico Kolter. 2018.
\newblock Provable defenses against adversarial examples via the convex outer
  adversarial polytope.
\newblock In \emph{Proceedings of the International Conference of Machine
  Learning}.

\bibitem[{Zellers et~al.(2018)Zellers, Bisk, Schwartz, and
  Choi}]{zellers2018swag}
Rowan Zellers, Yonatan Bisk, Roy Schwartz, and Yejin Choi. 2018.
\newblock {SWAG}: A large-scale adversarial dataset for grounded commonsense
  inference.
\newblock In \emph{Proceedings of Empirical Methods in Natural Language
  Processing}.

\end{thebibliography}

\end{document}